\documentclass[11pt,a4paper]{article}
\usepackage[hyperref]{acl2021}
\usepackage{times}
\usepackage{graphicx}
\usepackage{makecell}
\usepackage{floatrow}
\usepackage{subcaption}
\newfloatcommand{capbtabbox}{table}[][\FBwidth]
\usepackage{latexsym}

\newcommand{\boyuan}[1]{{\color{blue}#1}}

\usepackage{microtype}

\aclfinalcopy 

\title{Exploring Generalization Ability of Pretrained Language Models on Arithmetic and Logical Reasoning}

\author{Cunxiang Wang\textsuperscript{$\spadesuit$$\clubsuit$}, Boyuan Zheng\textsuperscript{$\clubsuit$$\diamondsuit$}, Yuchen Niu\textsuperscript{$\clubsuit$$\flat$} and Yue Zhang\textsuperscript{$\clubsuit$$\heartsuit$\thanks{\ \ The corresponding author} }\\
\textsuperscript{$\spadesuit$}Zhejiang University, China\\
\textsuperscript{$\clubsuit$}School of Engineering, Westlake University, China\\
\textsuperscript{$\heartsuit$}Institute of Advanced Technology, Westlake Institute for Advanced Study, China\\
\textsuperscript{$\diamondsuit$}Johns Hopkins University\\
\textsuperscript{$\flat$}Imperial College London\\
  {\tt \{wangcunxiang, zhangyue, zhengboyuan, niuyuchen\}@westlake.edu.cn} \\
  }

\date{}

\begin{document}
\maketitle
\begin{abstract}
To quantitatively and intuitively explore the generalization ability of pre-trained language models (PLMs), we have designed several tasks of arithmetic and logical reasoning. We both analyse how well PLMs generalize when the test data is in the same distribution as the train data and when it is different, for the latter analysis, we have also designed a cross-distribution test set other than the in-distribution test set. We conduct experiments on one of the most advanced and publicly released generative PLM - BART. Our research finds that the PLMs can easily generalize when the distribution is the same, however, it is still difficult for them to generalize out of the distribution.
\end{abstract}

\section{Introduction}

\begin{figure}[t]
  \centering{
  \includegraphics
  [width=5cm]  
  {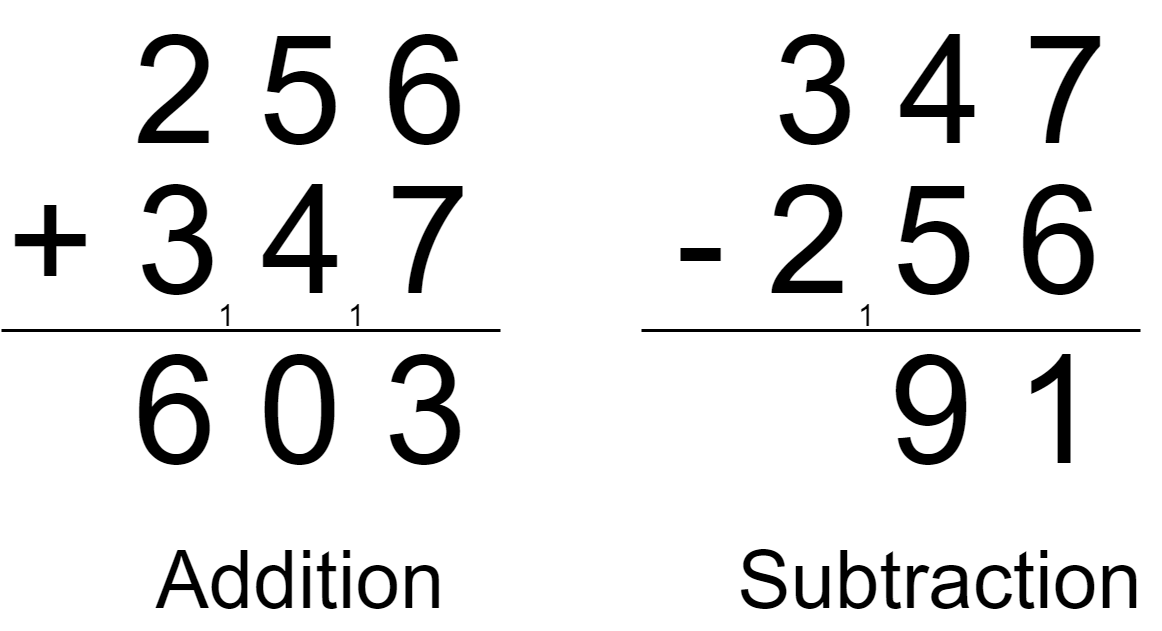}}
  \caption{Example mathematical rules for Addition and Subtraction. If the model can master these rules, we suppose it can generalize well on all two-number addition and subtraction samples. }
  \vspace{0mm}
  \label{rule}
\end{figure}

Neural networks have shown strong capabilities in a range of NLP tasks \citep{sutskever2014sequence,attention}. Recently, pretrained language models (PLMs) have achieved significantly levels of performance gains on many benchmark datasets  \citep{BERT,BART,GPT2}.
Recently, some work shows that neural networks are lack of generalization ability in mathematical and logical reasoning \citep{Nogueira2021InvestigatingTL,Madsen2019MeasuringAE}. This can lead to more understanding of the limitation of existing models and motivate future work. However, no work has been done to quantitatively or intuitively explore the conditions under which PLMs can generalize, in terms of whether PLMs can understand the internal mathematical rules and logical rules. The example of mathematical rules is shown in Figure~\ref{rule}. We suppose that if the model can effectively learn the underlying rules of Addition and Subtraction when giving sufficient training data, it can generalize to all two-number addition and subtraction calculation.

To this end, we conduct quantitative insights by designing a series of tasks for simple mathematical operations and logical reasoning, which includes numbering, addition, subtraction, comparison, and symbolic logic. We construct a set of corresponding datasets, where instances are in the form of text or  mathematical expressions. Some examples are shown in the next section. For example, in the Addition task, `100 + 200' is the question and `300' is the answer.

\vspace{0mm}

There are various types of generalization \citep{linzen-2020-accelerate,Lake2018GeneralizationWS}, such as question generalization on distribution differences  between training set and test set \citep{wallace-etal-2019-nlp}, and answer generalization on distribution differences between training set and test set \citep{Nogueira2021InvestigatingTL}. For example, in the Addition task, if the question and answer numbers in training data are of three-digit, but the question and answer numbers in the testing data are of two- or four-digit, they are in different distribution. 
To cover each type of generalization, we use different kinds of tasks and corresponding dataset. For example, we use \textit{addition} to test the generalization on the question distribution differences data between training and testing. In this task, the numbers in the training set and development have three digits. However, the numbers in test set is set to consist of two, three, and four digits.

We conduct experiments using BART \citep{BART} since they can generate arbitrary text sequences and have been shown to achieve the state-of-art results on numerous Natural Language Processing (NLP) tasks.
For each task, we fine-tune BART with training data, validate on the development set and finally evaluate on the test set. 
We find that strong PLMs can address simple generalization of the same answer distribution for counting, arithmetic and logic tasks. But they cannot master the underlying rules of arithmetic reasoning, for example, the model trained on 3-digit addition can handle the addition expressions with 2-digit or 4-digit. 

We will release all the code and data set for future study.

\section{Task}
We construct five tasks related to algebraic and logical reasoning, namely \textbf{Numbering, Addition, Subtraction, Comparison, Symbolic Logic}. 
In order to test the generalization ability of models on the data with the same distribution and on the data with the different distribution, we create an in-distribution dataset and a cross-distribution dataset for each task. The in-distribution dataset contains train set, development set and test set that are in the same distribution. The cross-distribution dataset only serves as the test set and it is in the different distribution in contrast to the in-distribution dataset. We believe that if the model can understand the underlying rules of arithmetic and logical Reasoning, it can both generalize well on in-distribution and cross-distribution test set.
\begin{table*}[!t]
  \centering
  \small
  \setlength{\tabcolsep}{1mm}
  \begin{tabular}{|c|c|c|c|c|c|c|}
  \hline
   \textbf{Task}&
  \makecell[c]{\textbf{Train Set}} & 
  \makecell[c]{\textbf{Dev Set}} & 
  \makecell[c]{\textbf{In- + Cross-} \\ \textbf{Distribution} \\ \textbf{Test Set}} & 
  \makecell[c]{\textbf{In- + Cross-} \\ \textbf{Distribution} \\\textbf{Dataset}}  \\
  \hline
  Numbering - Counting &  3,744  &  468 &  468 + 2,030 & 4,680 + 2,030\\
  \hline
  Numbering - Listing &  3,744  &  468 &  468& 4,680\\
  \hline
  Addition & 256,320   &  32,040 & 32,040 + 4,000& 320,400 + 4,000 \\
  \hline
  Subtraction & 256,320   &  32,040 & 32,040 + 4,000& 320,400 +  4,000\\
  \hline
  Comparison &  648,000  & 81,000  & 81,000 + 5,600 & 810,000 +  5,600 \\
  \hline
  \makecell[c]{Symbolic Logic} & 40,000 & 5,000 &  5,000 + 2,200 &  50,000 + 2,200 \\
  \hline
  \end{tabular}
  \label{dataset_details}
  \caption{Data statistics of each task. For each task, we list the in-distribution dataset and cross-distribution test set.
  }
\end{table*}

\subsection{Numbering}
\begin{figure}[t]
  \center{
  \includegraphics
  [width=7.5cm]  
  {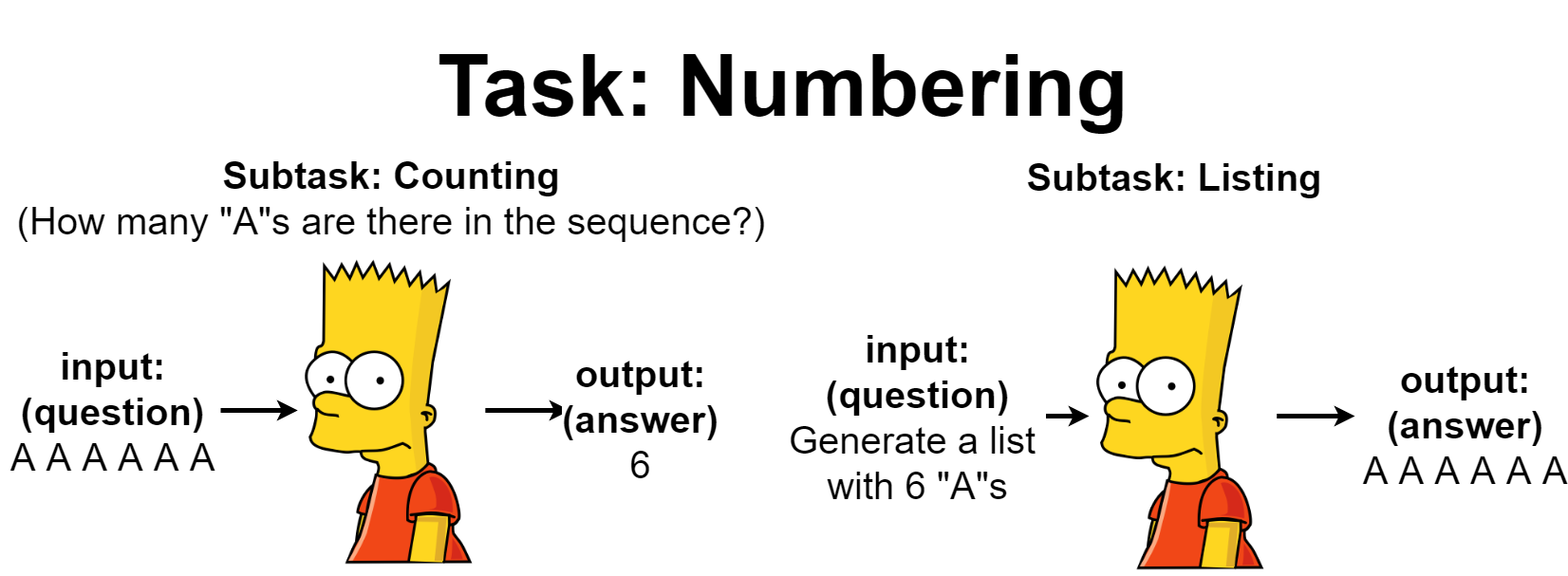}}
  \caption{The Numbering task has two subtasks, namely Counting and Listing.}
  \label{numbering}
  \vspace{0mm}
\end{figure}
This task comprises two symmetric subtasks, namely \textbf{Counting} and \textbf{Listing}.
Examples are shown in Figure~\ref{numbering}. The Counting task asks the model to count the number of characters in the input sequence. For example, `A A A A A A' is a sequence with length `6'. The Listing task asks the model to output a list with a specific length and character. For example, the model receives a command `Generate a list of 6 A' and the result is `A A A A A A'.

\subsection{Addition}
\label{sec:add}
\begin{figure}[t]
\centering
\begin{subfigure}{0.45\textwidth}
\includegraphics[width=0.9\textwidth]{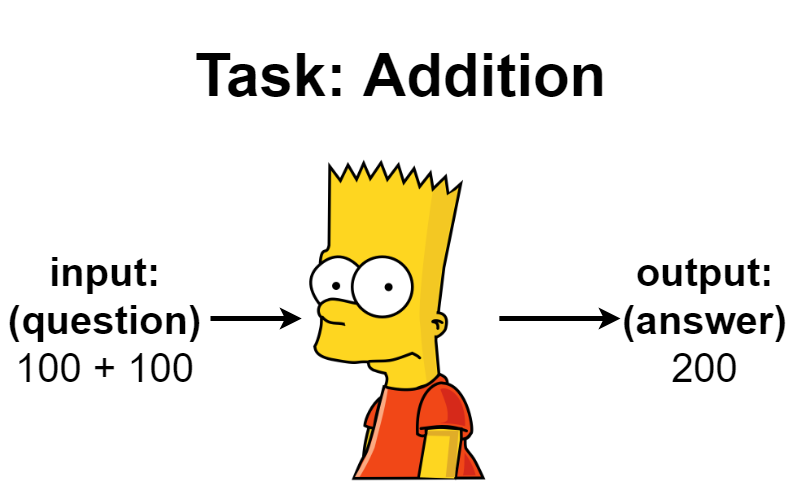}
\caption{Description of Addition task.}
\label{add}
\end{subfigure}
\hfill
\begin{subfigure}{0.45\textwidth}
\includegraphics[width=0.9\textwidth]{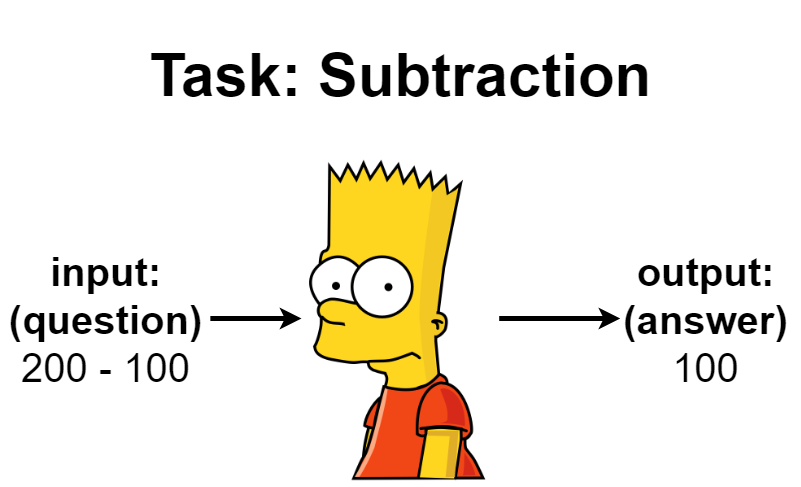}
\caption{Description of Subtraction task. }
\label{sub}
\end{subfigure}
\hfill
\begin{subfigure}{0.45\textwidth}
\includegraphics[width=0.9\textwidth]{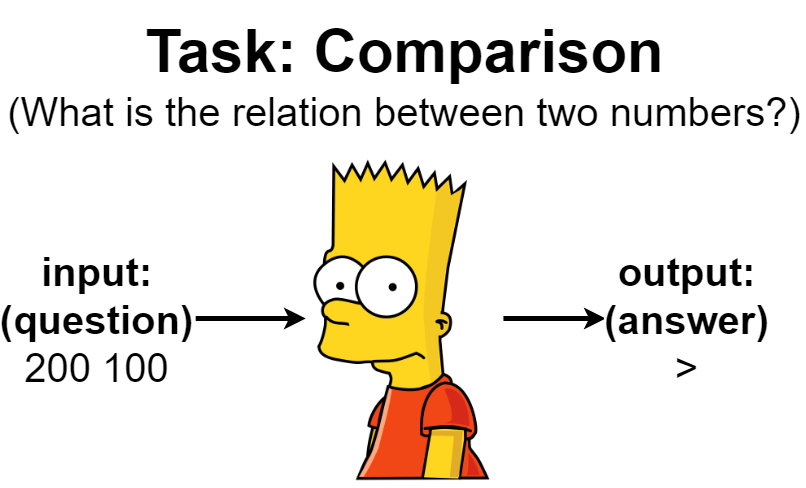}
\caption{Description of Comparison task.}
\label{compare}
\end{subfigure}
\hfill
\begin{subfigure}{0.45\textwidth}
\includegraphics[width=0.9\textwidth]{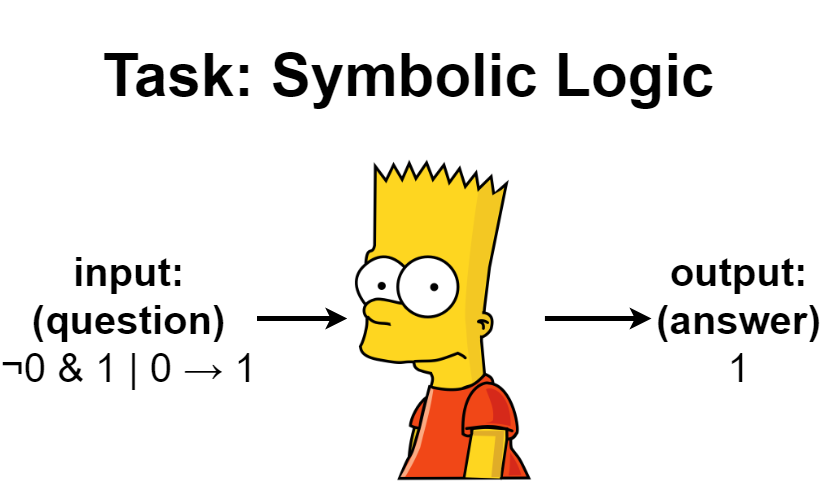}
\caption{Description of Symbolic Logic task. }
\label{logic}
\end{subfigure}
\vspace{0mm}
\end{figure}

The Addition task is the standard summation of two input numbers. In order to make sure that all numbers are in the same distribution during training, we use only the equations whose left-hand-side and right-hand-side are both \textit{three digits} in the in-distribution dataset. We also adopt two-digit and four-digit numbers on both sides in cross-distribution test set to further test the generalization ability of models.
One example is shown in Figure~\ref{add}. 

\subsection{Subtraction}

The Subtraction task the standard tack(?) to subtract a subtrahend from a minuend. In order to make sure that all numbers are in same distribution during training, we  use only equations whose left-hand-side and right-hand-side are both \textit{three digits} in the in-distribution dataset. We also adopt two-digit and four-digit numbers on both sides in cross-distribution test set to further test the generalization ability of models. A example of Subtraction task is shown in Figure~\ref{sub}.

\subsection{Comparison}


The Comparison task is to determine which of the two numbers is greater or smaller. In order to make sure all numbers are in same distribution during training, we use only equations whose left-hand-side and right-hand-side are both \textit{three digits} in the in-distribution dataset. We also adopt two-digit and four-digit numbers on both sides in cross-distribution test set to further test the generalization ability of models. One example is shown in Figure~\ref{compare}.

\subsection{Symbolic Logic}

As shown in Figure~\ref{logic}, this task is to reason over symbolic logic expressions. The input question expression consists of six basic components, which are `0', `1', `$\&$', `$|$', `$\neg$' and `$\to$', representing \textit{FALSE}, \textit{TRUE}, \textit{AND}, \textit{OR}, \textit{NOT} and \textit{IMPLY}, respectively. The output answer is either 0 or 1, which represent \textit{FALSE} and \textit{TRUE}, respectively. This task asks the model to reason over the input logic expression and determine whether it is true or false. 

In order to make sure all expressions are in the same distribution during training, we use only the expressions that contain \textbf{6 - 10} basic `0' and `1' components. For testing the generalization ability of models, we also adopt the some expressions with \textit{1 - 15} basic `0' and `1' components in the test set. 

Different from the other tasks, we select a subset from the overall dataset to serve as the in-distribution dataset because the data is large. We take only 10,000 of expressions with X basic components, where X is a number between 6 - 10, respectively. So, we end up with 50,000 samples in the in-distribution dataset.

\subsection{Metrics}
We use Exact Match to compute accuracy for Numbering, Addition, Subtraction and Comparison tasks. However, for the Symbolic Logic task, since the answer distribution is unbalanced (84\% answers are `1'), we use the F1 score as the metric.

\section{Experiments}
In this section, we separate the generalization experiments to \textbf{In-Distribution Generalization} experiments and \textbf{Cross-Distribution} experiments. In the former, the testing data is in the same distribution with the training data. In the latter, the testing data is in the different distribution from the training data. We suppose that if the model can master the underlying rules of the mathematical and logical reasoning, it should achieve 100\% accuracy on both In-Distribution Generalization experiments and Cross-Distribution experiments.

We have organized the details of in-distribution data and cross-distribution data in this section. In addition, We also sorted out the examples of them and put the examples in the  Appendix Table 1.

\subsection{Experimental Settings}
We adopt BART \citep{BART} namely due to the following reasons. First, it is a generative pretrained language model, which means that they can generate arbitrary sequences of tokens. This is essential for the addition and subtraction task. Second, it has achieved state-of-art results on numerous tasks and they has received much research attention. Last, it has released model checkpoints, thus it can be more standardized and more fair can evaluate them. 

For the BART \citep{BART} model, we conduct experiments on the publicly released `BART-Large' checkpoint \footnote{https://huggingface.co/facebook/bart-large/tree/main}.
We insert spaces between numbers while representing them in the data. For example, `111' is written as `1 1 1' both in the question and answer. For the character sequence in the Numbering task, we also insert spaces between the sequence, such as `A A A'.

\begin{figure}[t]
\centering
\begin{subfigure}{0.45\textwidth}
\includegraphics[width=\textwidth]{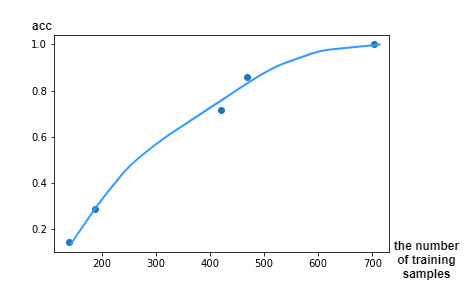}
\caption{The results of counting task.}
\label{list_results}
\end{subfigure}
\hfill
\begin{subfigure}{0.45\textwidth}
\includegraphics[width=\textwidth]{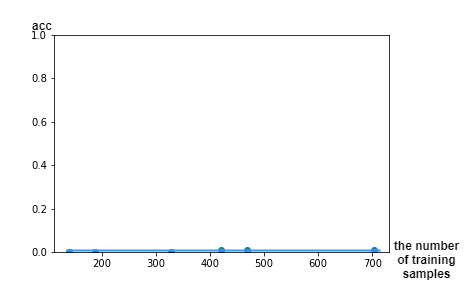}
\caption{The results of listing task.}
\label{count_results}
\end{subfigure}
\hfill
\begin{subfigure}{0.45\textwidth}
\includegraphics[width=\textwidth]{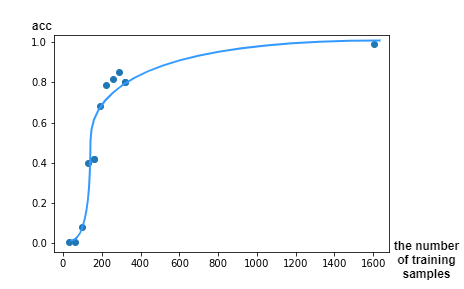}
\caption{The results of addition task.}
\label{add_results}
\end{subfigure}
\hfill
\begin{subfigure}{0.45\textwidth}
\includegraphics[width=\textwidth]{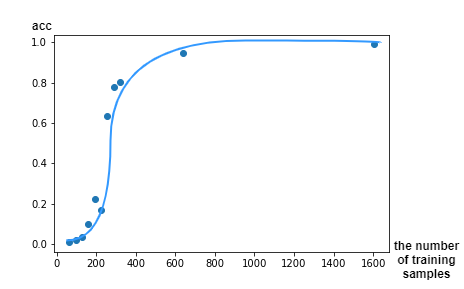}
\caption{The results of subtraction task.}
\label{sub_results}
\end{subfigure}
\hfill
\begin{subfigure}{0.45\textwidth}
\includegraphics[width=\textwidth]{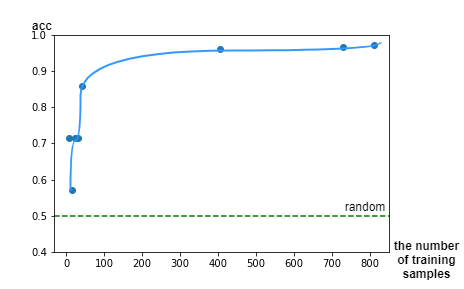}
\caption{The results of comparison task.}
\label{compare_results}
\end{subfigure}
\hfill
\begin{subfigure}{0.45\textwidth}
\includegraphics[width=\textwidth]{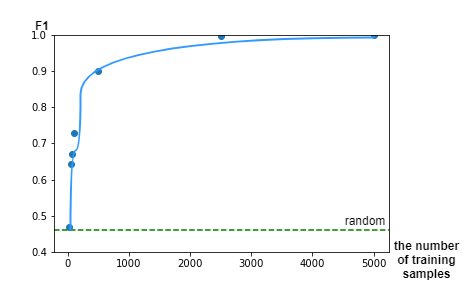}
\caption{The results of symbolic logic.}
\label{logic_results}
\end{subfigure}
\caption{The in-distribution results on each task.}
\end{figure}

\subsection{In-Distribution Generalization}

In this subsection, we mainly explore models' generalization ability on test data which in the same distribution with train data. 
For the Counting subtask of the Numbering task, each question is a sequence with 10-99 same character which is one character among the alphabet; each answer is an integer between 10 and 99.
For the Listing task, each question is a textual sequence `Generate a list with \textit{X Y}', where \textit{X} is an integer between 10 and 99 and \textit{Y} is one character among the alphabet; each answer is a sequence with 10-99 same characters.
For the Addition task, each question is an addition expression, and the answer is a sum number. Each number in the question and answer is three digits.
For the Subtraction task, each question is an subtraction expression 
, and the answer is a difference number. Each number in the question and answer is of three-digit. 
For the Comparison task, each question is made of two numbers and each answer is a single symbol which is either `$>$' or `$<$' or `$=$'. The numbers in the question are all of three-digit.
For the Symbolic Logic task, each question is a sequence with \textbf{5-10} basic `0' and `1' components; each answer is either 0 or 1.

For testing generalization ability on the same distributional data, we explore how the number of training samples affects the generalization. For each task, we extract subsets from the in-distribution train set and train on the subsets, but keep the distribution of development set and test set the same. Thus, we analyse how the number of training samples influences the performance, which also indicate the generalization ability of models on the data with the same distribution.

The in-distribution results on the Numbering task are shown in Figure~\ref{count_results} Figure~\ref{list_results}. 
For the Listing subtask, we find that the model's generation results are very unstable, which means that the outputs often contain other tokens other than the needed character. For example, when the input is `Generate a list of 6 A', the output can be `A A a Aa E T A'. When the sequence length increases, this kind of disruption will be more likely to occur. So, results are always around zero. We suppose this result is result from the instability of the generative model itself, because we also observe this situation from other generative models, such as T5 \citep{t5}. So, we mainly analyse the Counting task rather the Listing task in the following sections.

It can be seen that when the number of training samples increase, the performance of Counting will also improve.

The in-distribution results on the \textbf{Addition} task are shown in Figure~\ref{add_results}. We can seen that when the number of training samples is 1600 (0.5\% of the dataset), the model can achieve 99\% accuracy; even when the number of training samples is reduced to 160 (0.05\% of the dataset), the model can still achieve around 40\% accuracy. 
The in-distribution results on the \textbf{Subtraction}, \textbf{Comparison}, \textbf{Symbolic Logic} task are shown in Figure~\ref{sub_results}, Figure~\ref{compare_results} and Figure~\ref{logic_results}, respectively.
It can be seen from the figures that when the number of training samples increase, the model can perform better in the in-distribution test set. And when the training samples increase to several hundreds, the model can achieve around 100\% accuracy or F1, showing BART's ability on the in-distribution generalization.
Thus, we are wondering whether the model has truly learn the underlying rules of these tasks or they just use some spurious correlations to solve these questions, so, we design cross-distribution generalization test set to further explore the model's generalization ability in the following section.

\subsection{Cross-Distribution Generalization}

In this section, we analyse how models generalize (1) when test question distribution is different from train question while the test answer distribution is the same; (2) when test answer distribution is different from the train answer while the test question distribution keeps the same; (3) when the test question distribution and test answer distribution are both different from train set. We have designed testing data for different types of cross-distribution on each task and list examples of the testing data in this section.

\subsubsection{Varying Questions}
In this part, we mainly talk about when the test question distribution is different from the train question while the test answer distribution keeps the same, how strong is the model's generalization ability. So, we use the Counting, Addition, Subtraction, Comparison, and Symbolic Logic tasks to analyse.
For the Counting task, we use the instances whose character is not in letters of an alphabet while the number is still of two-digit. For example, the question is `$@\ @\ @\ @\ @\ @\ @\ @\ @\ @$' and the answer is `10'.
For the Addition task, we use the instances whose at least one added number is of two-digit. But we make sure answers of selected equations are all of three-digit. For example, the question is `50 + 170', the answer is `220'.
For the Subtraction task, the situation is similar to the Addition task, we use the instances whose at least one number is of four-digit. But we make sure answers of selected instances are all of three-digit. For example, the question is `1000 - 500', the answer is `500'.
For the Comparison task, the situation is also similar, we use the instances whose at least one number is of two-digit or four-digit. For example, the question is `56 176', the answer is `$<$'.
For the Symbolic Logic task, the situation is also similar, we use the instances which has 1 - 5 or 11 -15 basic `0' and `1' components. For example, the question is `$not\ 0\ and\ 1\ or\ 0$', the answer is `1'.

\subsubsection{Varying Answers}
In this part, we mainly talk about when the test answer distribution is different from the train answer while the test question distribution keeps the same, how strong is the model's generalization ability. As a result, we use the Addition and Subtraction to analyse.

For the Addition task, we use the instances whose two numbers are of three-digit while the answer is of four-digit. For example, the question is `500 + 600', the answer is `1100'.
For the Subtraction task, the situation is similar to the Addition task, we use the instances whose two numbers are of three-digit  while the answer is of two-digit. For example, the question is `550 - 500', the answer is `50'.

\subsubsection{Varying Instances}
In this part, we mainly talk about hen the test question distribution and test answer distribution are both different from the train set, how strong is the model's generalization ability. So, we use the Counting, Addition and Subtraction tasks to analyse.

For the Counting task, we use the instances whose character is not in letters of an alphabet and  number is not of two-digit. For example, the question is `$@\ @\ @\ @\ @\ @\ @\ @\ @$' and the answer is `9'.
For the Addition task, we use the instances whose at least one number in question is of two- or four-digit and the answer number is also of two- or four-digit. For example, the question is `50 + 960', the answer is `1010'.
For the Subtraction task, the situation is similar to the Addition task, we use the instances whose at least one number is of  two- or four-digit and the answer is also of two- or four-digit. For example, the question is `1100 - 50', the answer is `1050'.

\begin{table*}[]
\begin{tabular}{|c|c|c|c|}
\hline
Task & \makecell{Question \\ Cross-Distribution} & \makecell{Answer \\ Cross-Distribution} & \makecell{Instance \\ Cross-Distribution} \\
\hline
Counting       & 316/320(98.8\%)   &  /  & 0/1710     \\
\hline
Addition       & 15/1,500 (1.0\%) &  0/1,500 &   0/1,000  \\
\hline
Subtraction    &  13/1,500 (0.87\%) &  0/1,500  &  1/1,000 (0.1\%)   \\
\hline
Comparison     & 2,555/5,600(45.63\%)   &  /  &  / \\
\hline
Symbolic Logic & 2,200/2,200 (100\%)  &  /  &   /  \\ 
\hline
\end{tabular}
\caption{The performance of BART on cross-distribution test set. For each task and different distribution type, we select the model checkpoint which has achieved 100\% accuracy/F1 on the corresponding in-distribution test set. Note that the random result on Comparison is around 49.9\%. Data samples that models answer correctly on Addition and Subtraction task in the Cross-Distribution experiment can be found in Appendix~\ref{correct_samples}\boyuan{[Some of the examples here are deleted since they do not conform the fomat]}}
\label{cross_distribution_results}
\end{table*}


\subsubsection{Analysis on Different Cross-Distributions}
The model's performance on the test set of different types of cross-distributions is shown in Table~\ref{cross_distribution_results}. From the table, we can see that although BART has achieved 100\% accuracy on the in-distribution testing data, it fails to generalize on the cross-distribution testing data  of arithmetic reasoning tasks. 

Results of Counting and Symbolic Logic task on cross-distribution testing data are quite high.
However, for Counting task, all correct instances are the instances which have different length but have the same character distributions with the training data. In addition, the cross-distribution testing data only have length difference from the training data. 
Thus, we can conclude that the model is not sensitive to the length of question if the basic components does not change. 
This conclusion is also consistent with the result of \citep{ijcai2020-537}.
In addition, the results show that the model is especially weak in generalizing to the instances with different answer distributions.

To conclude, the model is still struggling on cross-distribution generalization, especially the carrying and borrowing in Addition and Subtraction tasks.
\subsection{Case Study on GPT-3}
GPT-3 \citep{GPT3} has received a lot of attention since it was born. And it has shown strong abilities on every single NLP task as well as on generalization. 
Thus, we also conduct some case study experiments of arithmetic calculation on GPT-3 \footnote{The experiments are conducted on https://beta.openai.com/examples/default-qa . But since the OpenAI has not released the whole API, we cannot finetune the model or do large scale experiments.}. We find that GPT-3 can handle the addition and subtraction calculation with 1,000 perfectly, but when the number increases, GPT-3 starts to lose its ability, it can only get some very specific instances correctly. A interesting case is that it can get correct result `9999999' from  `12345678 + 87654321', however, when we give it `12345678 + 8765432', it still answers `99999999'. We guess that the model does not have 
calculation ability, but rather remembers some examples that have appeared before, since each calculation with 1000 and `12345678 + 87654321' may appear in the Internet for many times while `12345678 + 8765432' may not so frequently appear.

\subsection{Overlap Analysis}

We have also explored how overlaps influence models' performance. 

Following \citep{overlap} and \citep{wang-etal-2021-generative},
if one test question appears in the questions of train set, we call it as \textbf{question overlap}, otherwise it is \textbf{question non-overlap}. Similarly, if one test answer appears in the answers of train set, we call it as \textbf{answer overlap}, otherwise it is \textbf{answer non-overlap}. If one test instance is both question overlap and answer overlap, we call it is \textbf{instance overlap}, otherwise it is \textbf{instance non-overlap}.

We mainly use results of the Addition task to illustrate this problem. 
However, for these two task, the question overlap is sightly different, if the two numbers of one test question both appear in the numbers of train set, we call it is \textbf{question overlap}, otherwise it is \textbf{question non-overlap}. Since one answer of these two tasks only contain one number, the situation is same with the original definition.

We choose the two results which using 1920 instances (0.6\% for the dataset) for training in the Addition task, because it has achieved 68\% accuracy, which means that the results have both correct and incorrect instances.

\begin{table*}
\footnotesize
\begin{minipage}[c]{0.33\textwidth}
\raggedright
\begin{tabular}{c|c|c}
\hline
& Correct & \makecell{Incorrect}\\
\hline
Overlap & 1,083 & 524 \\
\hline
\makecell{Non-\\Overlap} & 20,858 & 9,575\\
\hline
\end{tabular}
\subcaption{Instance Overlap}
\label{instance_overlap}
\end{minipage}
\begin{minipage}[c]{0.33\textwidth}
\begin{tabular}{c|c|c}
\hline
& Correct & \makecell{Incorrect}\\
\hline
Overlap & 4,538 & 1,846 \\
\hline
\makecell{Non-\\Overlap} & 12,403 & 8,253\\
\hline
\end{tabular}
\subcaption{question Overlap}
\label{question_overlap}
\end{minipage}%
\begin{minipage}[c]{0.33\textwidth}
\begin{tabular}{c|c|c}
\hline
& Correct & \makecell{Incorrect}\\
\hline
Overlap & 6,066 & 3,070 \\
\hline
\makecell{Non-\\Overlap} & 15,873 & 7,029\\
\hline 
\end{tabular}
\subcaption{answer Overlap}
\label{answer_overlap}
\end{minipage}

\caption{The overlap analysis.}
\label{Overlap}
\end{table*}

The results are shown in Table~\ref{Overlap}.
From the table, we can see that, unlike results from \citep{overlap} and \citep{wang-etal-2021-generative}, the overlap and non-overlap do not influence the models' performance. 

\section{Related Work}

Some works have investigated in Mathematical problems in NLP \citep{DROP,Math23L,Ape210K}. DROP \citep{DROP} is a reading comprehension dataset comprising several kinds of mathematical tasks, such as Subtraction and Selection. However, all answers of its questions can be directly or indirectly found in the corresponding passages. Math23L \citep{Math23L} is simple math word problem dataset with 23k problems. Its problem is of the simple English context format, along with the equation and the answer. Ape210K \citep{Ape210K} is a Chinese simple math word problem dataset with 210k questions. The questions are similar to Math23L's questions. The data are taken from some elementary school math word problems. These datasets do not contain a generalization test set, the test set is in the same distribution with the train set. In addition, the often used methods for these datasets are first to predict the equations or expression for the question and then to use calculation tool to get the result \citep{Math23L,Wangperawong2018AttendingTM}. However, our work concentrate on the generalization ability of models. Thus, we have designed test set with different distribution. In addition, we try to use the model to directly solve the questions, aiming test model's internal ability of understanding the deep rules of arithmetic and logical reasoning.

Some works have researched on models' the internal ability of solving mathematical expressions.  \citet{wallace-etal-2019-nlp} has investigated that how will different types of embedding, such as BERT \citep{BERT} and GloVe \citep{GloVe}, affect the performance of the same NAQANet model \citep{DROP} on the same tasks including List Maximum, Decoding and Addition. Besides, \citet{wallace-etal-2019-nlp} also explores that how the the way numbers are represented and the way to do tokenization affect the performance of models. \citet{geva-etal-2020-injecting} try to inject numerical reasoning skill by adding a calculation module into the PLMs, which helps the performance on DROP \citep{DROP} dataset.

There are also some works research focusing on the generalization ability of neural network models. \citet{Lake2018GeneralizationWS} research on the compositional generalization skills of sequence-to-sequence models, such as LSTM \citep{LSTM} and GRU \citep{GRU}. \citet{linzen-2020-accelerate} explain that the generalization test in machine learning (ML) is not very reasonable, they put forward seven suggestions to better evaluate the generalization ability of ML models.
\citet{overlap} and \citet{wang-etal-2021-generative} find that the PLMs cannot generalize well on Closed-book QA task \citep{how-much}, the model can handle the test instances which overlap with the train data, however, they cannot solve the non-overlapped instances. \citet{mccoy-etal-2020-berts} find that even when the model's architecture is set, the generalization ability of the model is still influenced largely by the random luck, the random initialized weights and other things. \citet{ijcai2020-537} perform Transformer-based models on simple logic reasoning test, and their results show that the model can get quite promising results and the model is not sensitive to the question length. Though \citet{Wang2020SemEval2020T4} proves that pretrained language models \citep{roberta, albert} can generalize well on textual commonsense reasoning tasks \citep{wang-etal-2019-make},  \citet{Wang2020CommonsenseKG} finds that transformer models \citep{comet} may not generalize well on commonsense knowledge graph \citep{atomic} reasoning. \citet{Zhang2020CanFP} analyses the generalization ability on the relation extraction task and find some specific problems can induce a significant decline in model performance. 

\section{Conclusion}
We have designed a series of tasks for evaluating BART on simple mathematical operations and logic reasoning, which includes numbering, addition, subtraction, comparison, and symbolic logic. We constructed a corresponding in-distribution datasets, and also designed cross-distribution test set to further evaluate the model's generalization ability. If the model can understand the underlying rules of these mathematical operations and logic reasoning, it can generalize well on both in-distribution and cross-distribution test set. Our experiments showed that BART can only generalize on the in-distribution test set but cannot perform well on  the cross-distribution test set, showing that the most advanced PLM still cannot understand the underlying rules of simple mathematical operations and logic reasoning.

\bibliographystyle{acl_natbib}
\bibliography{acl2021}


\appendix

\begin{table*}[]
\begin{tabular}{c|c|c|c|c}
\hline
Task &
  \begin{tabular}[c]{@{}c@{}}Training Data \&  \\ In-Distribution Test\end{tabular} &
  \multicolumn{3}{c}{Cross-Distribution Test} \\ \hline
 &
  &
  \begin{tabular}[c]{@{}c@{}} Question \\Cross- \\ Distribution\end{tabular} &
  \begin{tabular}[c]{@{}c@{}} Answer \\Cross- \\ Distribution\end{tabular} &
  \begin{tabular}[c]{@{}c@{}}Instance \\Cross- \\  Distribution\end{tabular} \\ \hline
Counting &
  \begin{tabular}[c]{@{}c@{}}a sequence\\  of {[}A-Z, a-z{]} \\ with a \\ length of \\ 10 to 99. \\ (B B B \\ B B B \\ B B B B)\end{tabular} &
  \begin{tabular}[c]{@{}c@{}}a sequence \\ of special \\ characters \\ with a \\ length of \\ 10 to 99\\ (@ @ @ \\ @ @ @ \\ @ @ @ @)\end{tabular} &
  &
  \begin{tabular}[c]{@{}c@{}}a sequence \\ of special \\ characters \\ with a \\ length of \\ 1 to 9 or \\ 100 to 1000. \\ (@ @)\end{tabular} \\ \hline
Addition &
  \begin{tabular}[c]{@{}c@{}}3-digit \\ addition.\\ (100 + 200 \\ = 300)\end{tabular} &
  \begin{tabular}[c]{@{}c@{}}at least \\ one addend \\ is 2 digits.\\ (50 + 170 \\ = 220)\end{tabular} &
  \begin{tabular}[c]{@{}c@{}}the answer \\ is 4 digits.\\ (500 + 600 \\ = 1100)\end{tabular} &
  \begin{tabular}[c]{@{}c@{}}at least one\\  number is \\ 2 or 4 digits.\\ (50 + 960 \\ = 1010)\end{tabular} \\ \hline
Subtraction &
  \begin{tabular}[c]{@{}c@{}}3-digit \\ subtraction.\\ (200 - 100 \\ = 100)\end{tabular} &
  \begin{tabular}[c]{@{}c@{}}at least \\ one number \\ is 4 digits,\\  but the answer \\ is still 3 digits.\\ (1000 - 500 \\ = 500)\end{tabular} &
  \begin{tabular}[c]{@{}c@{}}the answer \\ is 2 digits.\\ (550 - 500 \\ = 50)\end{tabular} &
  \begin{tabular}[c]{@{}c@{}}at least one \\ number is \\ 2 or 4 digits.\\ (1100 - 50 \\ = 1050)\end{tabular} \\ \hline
Comparison &
  \begin{tabular}[c]{@{}c@{}}3-digit \\ comparison.\\ (100 \textless 200)\end{tabular} &
  \begin{tabular}[c]{@{}c@{}}at least \\ one number \\ is not 3-digit.\\ (100 \textless 2000)\end{tabular} &
  &
  \\ \hline
\begin{tabular}[c]{@{}c@{}}Symbolic \\ Logic\end{tabular} &
  \begin{tabular}[c]{@{}c@{}}an equation \\ consists of \\ 6 to 10 \\ "0"s or "1"s.\\ (¬ 0 \& 1 | 0 \\ \& 1 | 1 | 0 is 1)\end{tabular} &
  \begin{tabular}[c]{@{}c@{}}an equation \\ consists of \\ 1 to 5 \\ or 11 to 15 \\ "0"s or "1"s.\\ (¬ 0 \& 1 | 0 is 1)\end{tabular} &
  &
  \\ \hline
\end{tabular}
\caption{Examples of training data, In-distribution test data, three kinds of cross-distribution tests. Note that the training data and in-distribution test share the same distribution.}
\label{cross_distribution_example}
\end{table*}

\section{Correct Cases of Cross-Distribution Experiments}
\label{correct_samples}
\subsection{Subtraction}
\label{correct_minus_sample}
\begin{tabular}{cr}
\hline
& 1101-974=127 \\
& 1070-955=115 \\
& 1069-959=110 \\
& 1111-991=120 \\
& 190-11=179 \\
& 222-99=123 \\
\hline
\end{tabular}

\subsection{Addition}
\label{correct_add_sample}
\begin{tabular}{cr}
\hline
& 75+653=728 \\
&77+849=926 \\
& 73+432=505 \\
& 42+668=710 \\
& 82+886=968 \\
& 80+874=954 \\

\hline
\end{tabular}

\end{document}